\documentclass[10pt, a4paper]{article}

\usepackage{graphicx}
\usepackage{tabularx}
\usepackage{soul}
\usepackage{microtype}
\usepackage{caption}
\usepackage{subcaption}
\usepackage{enumitem}
\setlist{nolistsep}
\usepackage{booktabs}
\usepackage{todonotes}
\usepackage{svg}
\usepackage{makecell}
\usepackage{siunitx} %
\usepackage{mathabx}
\usepackage{arydshln}

\usepackage{lrec-coling2024} 

\title{\textit{Willkommens-Merkel}, \textit{Chaos-Johnson}, and \textit{Tore-Klose}: Modeling\\ the Evaluative Meaning of German Personal Name Compounds}

\name{Annerose Eichel$^{1}$ {\bf \large Tana Deeg$^{1}$} {\bf \large André Blessing$^{1}$} {\bf \large Milena Belosevic$^{2}$}\\ {\bf \large Sabine Arndt-Lappe$^{3}$} {\bf \large Sabine Schulte im Walde$^{1}$}} 

\address{$^{1}$Institute for Natural Language Processing, University of Stuttgart\\$^{2}$ Faculty of Linguistics and Literary Studies, Department Linguistics, University of Bielefeld \\ $^{3}$English Linguistics and Trier Center for Language and Communication, Trier University \\
         \{annerose.eichel, tana.deeg, andre.blessing, schulte\}@ims.uni-stuttgart.de,\\
        milena.belosevic@uni-bielefeld.de, arndtlappe@uni-trier.de\\}

\abstract{
We present a comprehensive computational study of the under-investigated phenomenon of personal name compounds (PNCs) in German such as \textit{Willkommens-Merkel} (`Welcome-Merkel'). Prevalent in news, social media, and political discourse, PNCs are hypothesized to exhibit an evaluative function that is reflected in a more positive or negative perception as compared to the respective personal full name (such as Angela Merkel).
We model 321 PNCs and their corresponding full names at discourse level, and show that PNCs bear an evaluative nature that can be captured through a variety of computational methods. Specifically, we assess through valence information whether a PNC is more positively or negatively evaluative than the person's name, by applying and comparing two approaches using (i)~valence norms and (ii)~pretrained language models (PLMs). We further enrich our data with personal, domain-specific, and extra-linguistic information and perform a range of regression analyses revealing that factors including compound and modifier valence, domain, and political party membership influence how a PNC is evaluated.~

 \\ \newline }

\begin{document}

\maketitleabstract

\section{Introduction}

Personal name compounds (PNCs) such as \textit{Willkommens-Merkel} (`Welcome-Merkel'), \textit{Chaos-Johnson} (`Chaos-Johnson') and \textit{Tore-Klose} (`Goal-Klose') are nominal compounds that consist of a modifier such as \textit{Willkommen} (`Welcome') and a personal name such as \textit{Merkel}. PNCs are compositions that refer to a person, in our example the former German chancellor Angela Merkel. With few exceptions \cite{Wildgen:1981, Kürschner:2020}, PNCs have not received much attention from the theoretical or computational perspectives, but recent work suggests that they represent a rather frequent phenomenon
and carry an evaluative function with regard to the reference person \cite{belosevic:2021,Belosevic:2022}. That is, for a specific PNC in its discourse we hypothesize that the PNC is perceived as either more positively or as more negatively than the corresponding full name.
For understanding and generating texts from and for domains where real-world people are talked about (such as the news, social media, and any kind of political discourse, as well as for related tasks including sentiment and emotion analysis) it is thus particularly relevant to explore the evaluative nature and discourse effects of PNCs.

This paper performs such an investigation:
we leverage and extend an existing dataset consisting of German PNCs and their corresponding full names from the domains \textit{politics}, \textit{sports}, \textit{show business}, and \textit{others} \cite{belosevic:2021}. To assess PNCs in their contexts, we build a corpus drawing on data from social media (Twitter)
and German news (Deutscher Wortschatz).
We then draw on the notion of \textit{valence} from psycholinguistics that determines the pleasantness of a stimulus. Valence is considered one of the principal dimensions of affect and cognitive heuristics that shape human bias and attitude \cite{harmon2013}. It determines the affective quality referring to the intrinsic pleasantness or unpleasantness of a stimulus (e.g., \textit{joy} vs. \textit{toothache}) \cite{osgood1957measurement,frijda1986emotions}.
We hypothesize that PNCs with a higher or lower valence relative to their respective full name bear an evaluative character. To assess valence at context level, we develop two computational approaches.
First, we explore valence norms to efficiently compute and compare whether a PNC's contexts are more negative or positive than the contexts of the respective name. Second, we present an approach to interpret and evaluate the PNCs by leveraging a range of suitable pretrained language models (PLMs) that have been fine-tuned and evaluated for the conceptually related task of sentiment analysis 
\cite{barbieri-etal-2022-xlm,antypas-2022,guhr-etl-2020,lüdke-et-al-2022}. We use sentiment predictions as a proxy for valence and investigate whether PNCs for which more positive or negative sentiment is predicted relative to their respective full name bear an evaluative character. To this end, we compare results from four models varying regarding underlying architectures (RoBERTa, BERT) and training data.
Since PNC meaning is heavily dependent on modifier meaning, we also examine to which extent modifier valence influences the evaluation of the whole compound. 
Lastly, we explore whether personal background information such as age, domain-specific knowledge 
and extra-linguistics information
impact the PNC evaluation.
To explore which factors are influential at a statistically significant level, we
fit a range of regression models.

Our results show that PNCs are both positively and negatively evaluative in comparison to their full name with a tendency towards a negative evaluative nature, underlining previous findings from \citet{belosevic:2021, Belosevic:2022}. We find domain-specific differences with public figures from the domain \textit{politics} bearing a more negative meaning, while the opposite is true for PNCs from the domain \textit{sports} and \textit{show business}.
Modeling modifiers reveals corresponding valence scores to be more extreme than valence scores obtained for the compound as a whole with a tendency towards lower valence. Our findings also highlight cases where modifier meaning is either interpreted non-literally or smoothed down when evaluated as PNC constituent. Furthermore, comparing results across approaches shows that valence assessments using PLMs lead to up to 37\% more negatively evaluated PNCs as compared to results based on valence norms.
Finally, while personal and domain-specific information impact the evaluative nature of a PNC, regression models including extra-linguistic information such as compound valence are more informative. Our best model thus combines information from all variables except name valence, with factors such as compound valence, domain, and political party membership playing an important role.

\section{Background and Related Work}

\subsection{Personal Name Compounds (PNCs)}
PNCs such as \textit{Tore-Klose} (`Goal-Klose') are nominal compounds that consist of a modifier which is usually realized in form of appellative or onymic constituents (e.g., \textit{Tore)} and a  head constituent that is filled with a first, last, or nick name (\textit{Klose}) \cite{Belosevic:2022}.
PNCs are formed based on regular patterns within a context that both evaluates and evokes knowledge regarding the name bearer \cite{belosevic:2021}. For example, the PNC \textit{Tore-Klose} ('Goal Klose') refers to the former German soccer player Miroslav Klose who is the all-time top scorer for Germany with 16 goals scored during the Men's FIFA World Cup. The example also illustrates the importance of the compound modifier contributing information regarding the name bearer or events in which the name bearer was involved. In other words, the meaning of the modifier is the reason or at least related to the reason why this compound was formed. In our example, \textit{goal} hints towards a positive evaluation of the PNC as such an event is usually connected with particular athletic performance and special occasions, as well as concrete events such as scored goals during the soccer world cup in 2014. 

\subsection{Approaches to Modelling PNCs}
German PNCs are under-investigated from both a theoretical and computational point of view. Early work is limited to very small scale studies based on a few names \cite{Wildgen:1981} or focus on other phenomena and touch on this composition type only in passing \cite{Kürschner:2020, ortner:1984,ortner:1991,Schlücker:2017,Schlücker:2020}. An exception is recent work by \citet{belosevic:2021} who present a systematic analysis of \textasciitilde1.2K PNCs to test three hypotheses on personal name composition that prevail in word formation (irregularity or unpredictability, low frequency, evaluative function). They compile a small Twitter and newspaper corpus, manually infer a paraphrase of the PNC in form of a relative clause, and assign a corresponding \href{https://gsw.phil.hhu.de/}{German FrameNet} \cite{Ziem2014} relation. Their corpus analysis shows that not only are PNCs formed based on regular patterns but also bear an evaluative and a knowledge-evoking function.
While this analysis constitutes an important contribution to name-based composition and evaluation, it is, however, limited by size and a manual approach. 

From a NLP perspective, PNCs have not received much attention yet. Related tasks such as noun compound interpretation where a noun compound is classified into a predefined label or expressed in a 
paraphrase \cite{lauer-1995-corpus,kimAndBaldwin:2005,shwartz-dagan-2018-paraphrase,coil-shwartz-2023-chocolate} and noun compound conceptualization exploring rare or novel interpretation through paraphrasing \cite{dhar-etal-2019-measuring,li-etal-2022-systematicity} neither include PNCs nor approaches to assess the evaluative function of such compounds. 
Another relevant line of work concentrates on sentiment analysis (SA), i.e., predicting the sentiment, attitude or opinion of text or speech on different units using e.g., the categories positive, neutral and negative \cite{mohammad-2012-emotional}.
\begin{table*}[!htpb]
\small
\begin{tabularx}{\textwidth}{ lX }
\toprule
\textbf{PNC}               & \textbf{Context}                                                                                                                                          \\ \midrule
\parbox[t]{4cm} {\textit{Willkommens-Merkel} \\ (`Welcome-Merkel')} & \#Lanz This political constellation should never have come about in the first place, says Merz. Another declaration of war on \textit{Welcome-Merkel} \\
\midrule
\parbox[t]{4cm} {\textit{Villen-Spahn} \\ (`Villas-Spahn')} & I'm so fed up with jet-setting, fizzy brew-drinking politicians who flaunt their swagger. Where do they get all the money from? I would like to see more transparency in the revenues of \textit{Villas-Spahn} and \textit{Jet-Merz}, for example. \\ \midrule
\parbox[t]{4cm} {\textit{Gedächtnislücken-Scholz} \\ (`Memory-Lapse-Scholz')} & Do I understand correctly that the same \textit{Memory-Lapse-Scholz}, who seems to have lost all decency in connection with the huge tax fraud Cum-Ex, is now hypocritically demanding -- morality? Morality? Scholz? Really? 
\\ \midrule
\parbox[t]{4cm} {\textit{Vollgas-Vettel} \\ (`Pedal-to-the-Metal-Vettel')} & Excellent! - \textit{``Pedal-to-the-Metal-Vettel''} saves World Cup lead \#Vettel                                                                            \\  \midrule 
\parbox[t]{4cm} {\textit{Gold-Rosi} \\ (`Gold-Rosi')} & 
Ski legend Rosi Mittermaier has died at the age of 72. \textit{``Gold-Rosi''} became the ``pop star'' of winter sports at the 1976 Winter Olympics. \#rosimittermaier \\ \bottomrule
\end{tabularx}
\caption{Sample PNCs (marked in italics) from the domain \textit{politics} (Merkel, Scholz, Spahn) and \textit{sports} (Vettel, Rosi) in context (translated from German). }
\label{tab:data-example}
\vspace{-0.3cm}
\end{table*}
While an increasing amount of researchers have explored the sentiment of news text and tweets in many languages including German \cite{cieliebak-etal-2017-twitter, fehle-etal-2021-lexicon,grimminger-klinger-2021-hate,schmidt-etal-2022-sentiment,zielinski-etal-2023-dataset}, 
PNCs have not been investigated yet. We address this gap by developing two computational approaches drawing on valence norms and PLMs fine-tuned for SA to examine the evaluative nature of PNCs at discourse level.

\section{Data}

\subsection{Targets}

\paragraph{Personal Name Compounds (PNCs)}
We start out with 770 eventive PNCs provided by \citet{Belosevic:2022}. As described in detail in her work, the PNCs were collected manually by searching for the string *name or -name as well as regular expressions in the DWDS \citep{goldhahn-etal-2012-building} WebXL interface. Additional targets were collected via the Twitter Extended Search option.
We filter the corpus for PNCs with a common or proper noun modifier followed by a personal name such as \textit{Willkommens-Merkel} or \textit{Gold-Rosi}, and only keep PNCs for which we find a context instance (cf. §\ref{subsec:context_corpora}).
This leads to final lists of 321 and 217 instances of PNCs with at least one and five contexts that are used for modeling, respectively. To maximize recall w.r.t our corpora, PNCs are modified at character level using heuristics such as eszett replacement: {\ss} $\rightarrow$ ss, e.g., \textit{Spa{\ss}-Guido} $\rightarrow$ Spass-Guido ('fun-Guido).\footnote{See App.~\ref{appsec:data} for the full list of heuristics.}

The PNCs can be categorized into the domains \textit{politics} including politicians such as Angela Merkel and Boris Johnson (87\%), \textit{sports} referring to athletes who are mostly soccer players but also include e.g., the Formula 1 driver Sebastian Vettel (9\%), \textit{show business} encompassing e.g., the actress Angelina Jolie (1\%), and \textit{others} including public figures such as lobbyist Karlheinz Schreiber or the climate activist Greta Thunberg (3\%). We list example PNCs
within a context in Table~\ref{tab:data-example} and make the full list of examined PNCs publicly available.\footnote{The full list is available here: \url{https://github.com/AnneroseEichel/LREC-COLING2024}}

\paragraph{Full Names}
We manually map each PNC to the corresponding full name (first name, last name), yielding 131 and 113 names for which at least one or five PNC instances are in the used corpora.

\subsection{Context Corpora} \label{subsec:context_corpora}
For our models, we build two corpora based on Twitter and Deutscher Wortschatz. Each corpus consists of two subcorpora containing contexts for full target names or PNCs.

\paragraph{Twitter}
We download tweets containing PNCs or full names using the Twitter\footnote{We downloaded the data before the re-naming.} 
Academic API (closed in spring 2023) using \href{https://twarc-project.readthedocs.io/en/latest/twarc2_en_us/}\texttt{twarc}. Modifiers and names are required to be perfect matches while characters in between are not restricted to hyphens or whitespace but may also be e.g., hashtags. The maximum context is defined as one tweet and added to the corresponding subcorpus whenever a PNC or name is found. For full names, we download 100 tweets per match and remove retweets based on URLs.
This yields a number of 9,145 and 24,688 tweets containing a PNC or full name, respectively.

\begin{figure*}[!htpb]
    \centering
    \includegraphics[width=\textwidth]{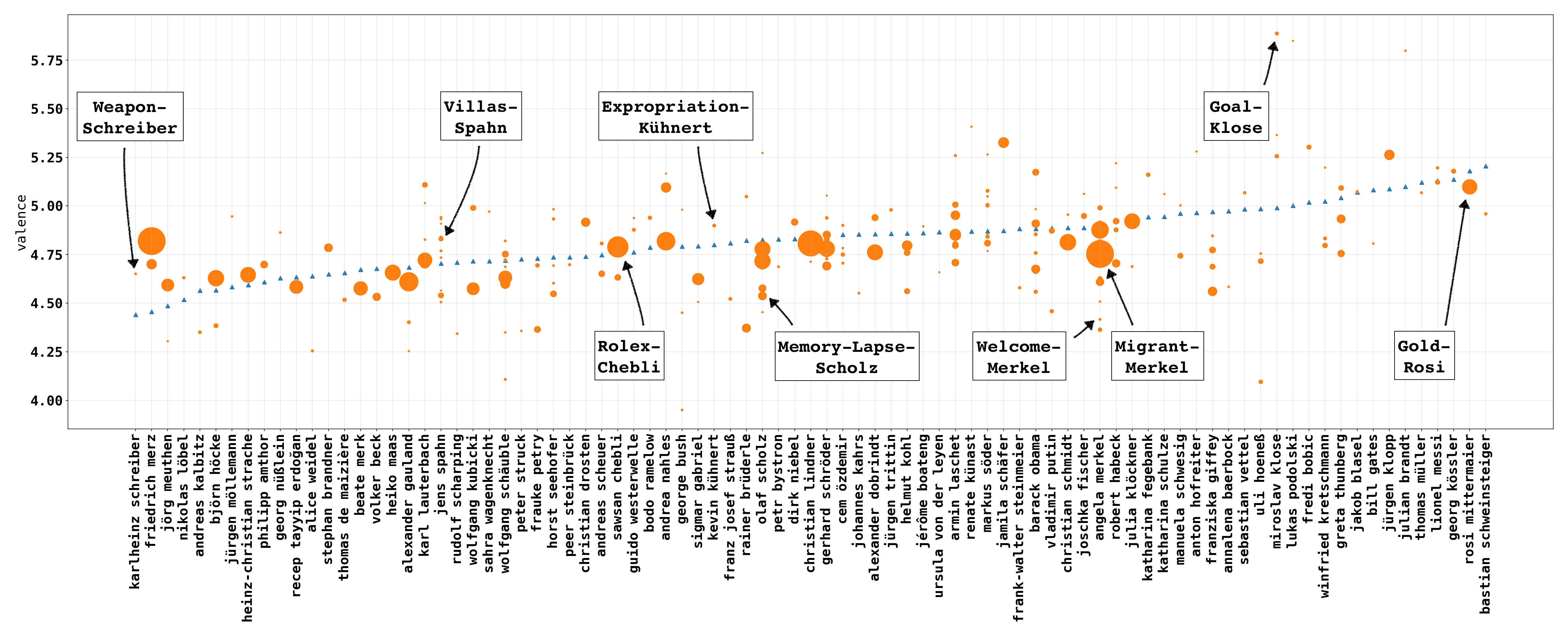}
    \caption{Overview of name (blue triangles) vs. PNC (orange dots) valence with PNC frequency visualized by size of orange dots (minimum frequency = 5). One ore more PNCs can relate to one name, e.g., \textit{Willkommens-Merkel }(`Welcome-Merkel') and \textit{Migranten-Merkel} (`Migrant-Merkel') both referring to Angela Merkel  and bearing a more negatively evaluative character than the name itself.}
    \vspace{-0.3cm}
    \label{fig:pnc-name}
\end{figure*}

\paragraph{Deutscher Wortschatz} 
We further leverage the Leipzig Corpora Collection providing large numbers of German news data in the context of the ongoing project \textit{Deutscher Wortschatz} (DW) \citep{klein-geyken-2010,goldhahn-etal-2012-building}. We leverage the full DW data inventory of \textasciitilde27 million sentences. We define a context as a sentence that we only add whenever they contain a PNC or a name.
This yields a total number of 170 and 233,477 sentences containing a PNC or full name, respectively.

\section{Evaluating PNCs via Valence} \label{sec:valence}

As a first step, we explore the evaluative nature of PNCs from a range of domains drawing on the notion of \textit{valence} from psycholinguistics, determining the pleasantness of a stimulus (\textit{joy} vs. \textit{toothache}).
We hypothesize that PNCs with higher or lower valence relative to their respective full name bear an evaluative character. For this, PNC and full name valence are assessed and compared at context level both cross and within domains. We further determine PNC modifier valence and explore the relationship between PNC and modifier evaluation. 

\subsection{Assessing Context-Level Valence}

\paragraph{Valence Norms} 
We use the automatically generated valence norms by \citet{koper-schulte-im-walde-2016-automatically} who provide ratings on a scale from 0 to 10 with 0 and 10 referring to low and high valence, respectively. Provided norm types are lower-cased and provided in their inflected forms.

\paragraph{Valence Exploration}
We apply part-of-speech (PoS) tagging\footnote{We use the \texttt{TreeTagger} \cite{Schmid1999} which we find to produce better results for the task and text at hand than more recent libraries, e.g., \texttt{spaCy}.} including lemmatization to all context words of a given target. We only keep context lemmas which belong to the word classes noun, adjective, or verb.\footnote{See App.~\ref{appsec:data} for the full list of PoS tags.} 
Then, each lemma is assigned a valence score using the valence norms by \citet{koper-schulte-im-walde-2016-automatically}, iff available. 
Specifically, the valence of a target $(t)$ is defined by the normalized mean valence of the corresponding sum of context lemmas $(W_{t})$:
\begin{equation}
    valence(t) = \frac{1}{|W_{t}|} * \sum_{w \in W_{t}} valence(w)
\end{equation}
We compare PNC and name valence by means of the valence delta $\Delta$ both across and within the domains \textit{politics}, \textit{sports}, and \textit{others}, and determine statistical significance by calculating the Pearson correlation coefficient.

\paragraph{Cross-Domain Results} Our findings are visualized in Fig.~\ref{fig:pnc-name} including PNC frequency illustrated by increasing dot size. PNCs are sorted by name valence with the lowest valence score determined for the German businessman, arms dealer, and lobbyist \textit{Karlheinz Schreiber},
and the highest valence score calculated for the former German soccer player \textit{Bastian Schweinsteiger}. We note that PNC valence moves more towards the name valence line in case of higher frequencies, while outlier PNC scores tend to be more distanced. 
Across domains, we find PNCs bearing a slightly more negative nature than full names with PNC valence distributed over a greater score range. More specifically, 56\% of PNCs are shown to be more negatively evaluative than the corresponding full names.
The Spearman correlation coefficient reveals a moderately positive correlation of statistical significance ($\rho=0.43$, $p<0.01$).

To gain qualitative insights, we inspect the sample name-PNC pairs with the largest positive and negative relative differences $\Delta$ and examine the most frequent context words.
The greatest positive $\Delta$ 0.9 comes from the PNC \textit{Tore-Klose} (`Goal-Klose') which refers to the former German soccer player Miroslav Klose. 
While name valence is around average (4.99), the PNC is evaluated very positively (5.89). Frequent context words of both PNC and name are on average positive since Klose was a very successful athlete, known for his fair play, and well-received in the public sphere. However, considering that the PNC \textit{Tore-Klose} refers to the specific positive event of scoring many goals, the PNC is evaluated even more positively, with frequent context words such as \textit{feiern} (`celebrate'), \textit{herrlich} (`wonderful'), and \textit{gold} (`gold') pointing at this finding. 
When looking at the pair with the greatest negative difference, we find the PNC \textit{Knast-Hoeneß} (`Prison-Hoeneß') with a $\Delta$ of -0.89.
While Ulrich Hoeneß is also a successful former German soccer player, he is now mainly known for the fact that he was found guilty for serious cases of tax evasion. However, frequent context words are mixed, including the modifier \textit{Knast} (`prison'), \textit{sagen} (`to say'), and \textit{Jahr} (`year').
An explanation for this might be that the public credits Hoeneß for accepting the guilty verdict without appeal.

\begin{figure}[!htpb]
    \centering
    \includegraphics[width=0.45\textwidth]{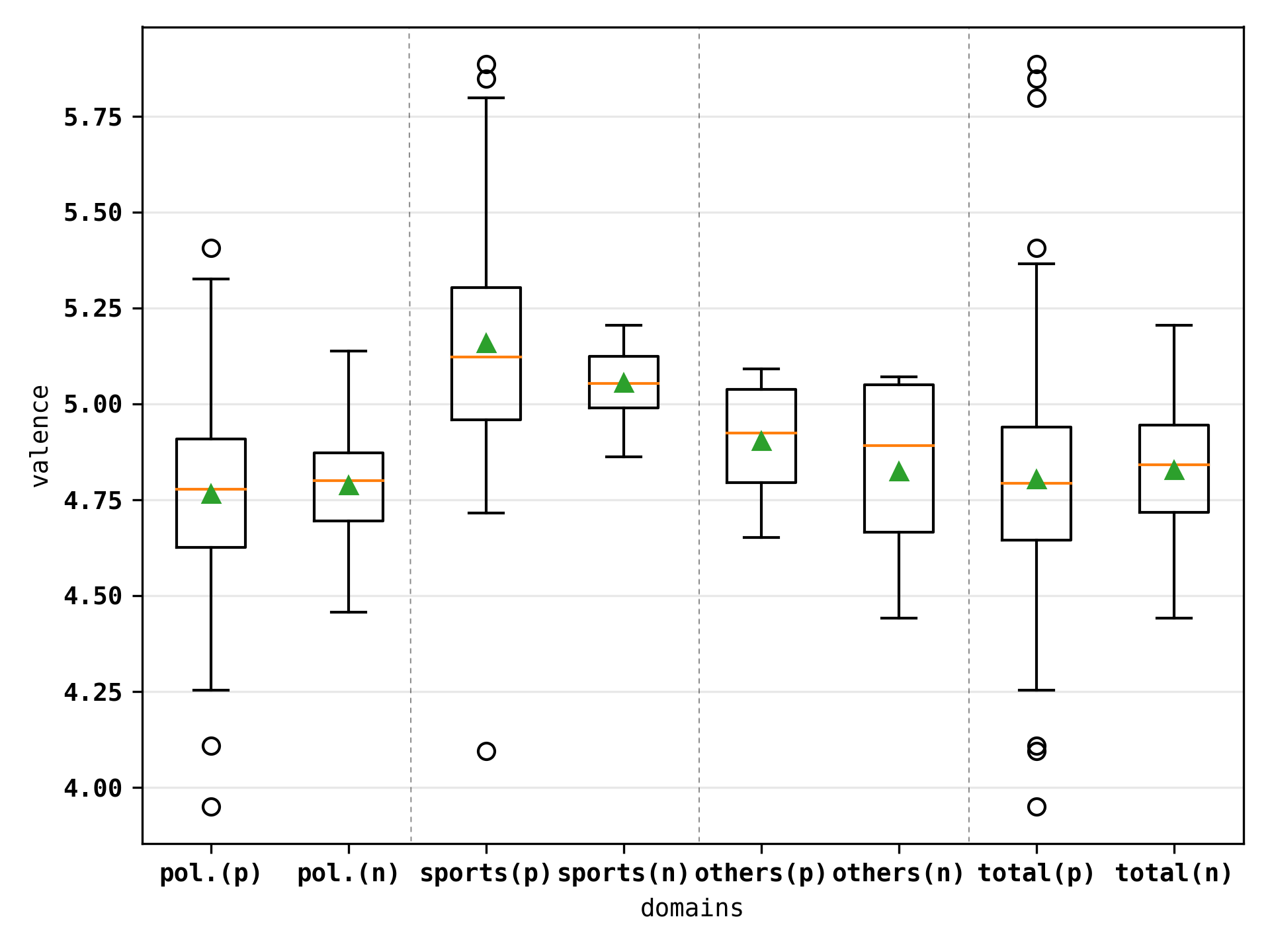}
    \caption{Domain-specific (\textit{politics}: pol, \textit{sports}, \textit{others}) and cross-domain valence comparison for PNCs (p) and names (n). Green triangles and orange lines illustrate arithmetic mean and median values, respectively. Min.\ PNC freq.\ = 5.}
    \label{fig:domain-specific-comparison}
    \vspace{-0.2cm}
\end{figure}

\paragraph{Domain-Specific Results} We further compare results within domains. As shown in Fig.~\ref{fig:domain-specific-comparison}, PNC valence scores extend both below and above name valence for the domains \textit{politics} and \textit{sports}, while exceeding name valence only for \textit{others}. Both PNC and name valence scores are lowest for public figures from \textit{politics}, followed by the slightly more positively evaluated domain \textit{others}. Athletes, in contrast, are generally evaluated more positively, surpassing the mean value of 5 for both PNC and name valence. 

\subsection{Assessing Modifier Valence}

PNCs constitute determinative compounds where a modifier such as a noun modifies the compound head, in our case, a name. As modifier meaning has a large share in human compound interpretation, we would assume that modifier meaning influences the way the whole compound is evaluated.
We thus hypothesize that modifier meaning can be used as a proxy for compound evaluation. For this, we examine the connection between PNC and modifier valence.
We manually\footnote{Results using libraries such as \texttt{TreeTagger} or \texttt{spaCy} yield very inaccurate results which could not be used for further investigations.} determine modifier lemmas and automatically assign a valence score \cite{koper-schulte-im-walde-2016-automatically} whenever possible. \footnote{In the rare case of double entries, we choose among the available scores at random.}
We calculate the relative difference $\Delta$ between the 203 PNC and name valence scores and determine statistical significance using Spearman's $\rho$.

\paragraph{Cross-Domain Results} We find modifier valence to be spread across a substantially wider range than PNC valence, with minimum modifier valence as low as 0.89 (\textit{folter}: \textit{Folter-Bush} (`Torture Bush')) and maximum modifier valence going up to 7.9 (\textit{willkommen}: \textit{Willkommens-Merkel} (`Welcome-Merkel')). In comparison, PNC valence values range between 3.95 (\textit{Folter-Bush} (`Torture-Bush')) and 5.89 (\textit{Tore-Klose} (`Goal-Klose')), with average PNC valence at 4.81 and average modifier valence at 4.22. The majority of modifiers is located below PNC valence with modifier valence increasing only slightly with higher PNC valence (Spearman's $\rho=0.50$, $p<0.01$).

For more fine-grained insights, we examine the modifier-PNC pairs with the largest positive and negative difference $\Delta$. Investigating the largest positive $\Delta=3.48$ leads us to the PNC \textit{Willkommens-Merkel} (`Welcome-Merkel'), 
with peak modifier valence (7.9) and below-average PNC valence (4.42). Inspecting frequent context words such as \textit{Abschiebung} (`deportation'), \textit{verheerend} (`devastating'), and \textit{Kritik} (`criticism') reveals quite negative discourses. This might indicate potentially ironic use of the modifier \textit{welcome} as the context words convey the opposite meaning or a negative stance towards corresponding political actions of the name bearer.
The modifier-PNC pair \textit{Enteignungs-Kühnert} (`Expropriation-Kühnert') is the PNC with the largest negative $\Delta=-3.89$. Here, modifier valence is extremely low (1.51), while PNC valence is around average (4.9). Context words are very mixed w.r.t. valence, including mentions of \textit{Partei} (`political party'), \textit{Wahl} (`election'), and \textit{Wohnung} (`apartment'). Thus, the extreme value of the modifier is not reflected by the context words and, consequently, PNC valence.

\section{PLMs for Evaluating PNCs} \label{sec:plms}

We further explore the evaluative function of PNCs leveraging a range of suitable pretrained language models (PLMs) that have been fine-tuned and evaluated for the task of sentiment analysis (SA).
We propose to use sentiment predictions as a proxy for valence and hypothesize that PNCs for which more positive or negative sentiment relative to their respective full name is predicted to bear an evaluative character. For this, we formulate the task of predicting the evaluative nature of a PNC in context as a text classification problem at the document level. We feed the context including a PNC or name to a model and obtain top-1 predictions. The multi-class output is then mapped onto a valence scale to calculate relative differences between PNC and name valence that are leveraged as a proxy for the evaluative nature of a PNC. Results are compared across different models with varying underlying architectures and training data as well as findings from experiments based on valence norms (§\ref{sec:valence}). 

\paragraph{Models} \citet{barbieri-etal-2022-xlm} devise a multilingual XLM-RoBERTa model (\href{huggingface.co/cardiffnlp/twitter-xlm-roberta-base-sentiment}{XLM-Twitter}) trained on ~198M tweets and fine-tuned on SA for 8 languages including German. \citet{antypas-2022}  harness this model and provide a version that is further fine-tuned on sentiment by politician's tweets, focusing on MPs from the UK, Spain, and Greece (\href{huggingface.co/cardiffnlp/xlm-twitter-politics-sentiment}{XLM-Politics}). Although no explicit fine-tuning has been performed for German, we hypothesize that parameter changes may still yield interesting changes for PNCs referring to politicians. We further test a model specifically focusing on German \cite{guhr-etl-2020} which is based on the German BERT architecture and trained on 1.834M German language samples from domains such as Twitter, Facebook, and reviews (\href{huggingface.co/oliverguhr/german-sentiment-bert}{GBERT-Sentiment}). We also explore a model extending \citet{guhr-etl-2020}'s model by additional fine-tuning on German news texts about migration \cite{lüdke-et-al-2022} (\href{huggingface.co/mdraw/german-news-sentiment-bert}{GBERT-Migration}).

\paragraph{Experimental Setup} We feed the context including a PNC or name to a model and obtain top-1 predictions with a label $l$ where $l\in$ \{negative, neutral, positive\}, respectively.\footnote{All experiments are performed with one NVIDIA RTX A6000 GPU with inference runtime per model at max. 4 minutes.} 
To map the obtained labels onto our valence scale, we compute a weighted valence score for each target $t$ (PNC or name) with

\vspace{-0.15cm}
\begin{equation} \label{eq:2}
    valence(t) = 
    \frac{\sum\limits_{l_{pos} \in L_{t}} + 0.5 * \sum\limits_{l_{neu} \in L_{t}}}{L_{t}} * 10
    \vspace{-0.15cm}
\end{equation}
where $l_{pos}$ and $l_{neu}$ denote \textit{positive} and \textit{neutral} labels obtained for $t$, and $L_{t}$ refers to the sum of labels observed for $t$. In principle, valence could also be defined as (i)~the sum of all \textit{positive} labels only, or (ii)~$1-$ sum of all \textit{negative} labels, normalized by the sum of all labels. However, (i) does not include the overall label distribution,
and (ii) sums all \textit{positive} and \textit{neutral} labels as \textit{positive} labels. In contrast, our approach incorporates whether the remaining labels are mainly \textit{neutral} or \textit{negative}, while weighing contributions of \textit{positive} and \textit{neutral} differently.

\paragraph{Cross-Domain Results}
Results from our PLM experiments suggest that PNCs carry a clearly more negative evaluative function than full names across all tested models with up to 93.55\% PNCs labeled as more negatively evaluative than the corresponding  name (cf. Table~\ref{tab:model-overview}). We find low positive correlations of statistical significance between name and PNC valence for the XLM-RoBERTa-based models and no significant correlation for the BERT-based models focusing on German data. 

When comparing results to our valence experiments (§\ref{sec:valence}), the largest difference can be seen in cases where PLMs predict a PNC to be more negatively evaluative than the full name (XLM-Twitter: 36\%, XLM-Politics: 31\%, GBERT-Sentiment: 39\%, GBERT-Migration: 35\%, cf. Table~\ref{tab:approach-comparison}). 

\begin{table}[!thpb]
\centering
\begin{tabular}{@{}l|rr@{}}
\toprule
                & \multicolumn{1}{l}{$\Delta<0$} & \multicolumn{1}{l}{$\Delta>0$} \\ \midrule
XLM-Twitter     & 90.32                            & 9.68                                \\
XLM-Politics    & 82.49                            & 17.51                               \\
GBERT-Sentiment & 93.55                            & 6.45                                \\
GBERT-Migration & 89.86                           & 10.14                              \\ \bottomrule
\end{tabular}
\caption{Overview of relative difference values ($\Delta$) between PNC and name valence with $\Delta<0$ referring to the PNC bearing a more negative meaning and $\Delta>0$ suggesting the PNC to be more positively perceived than the respective name.}
\label{tab:model-overview}
\vspace{-0.4cm}
\end{table}

\begin{table}[!htpb]
\centering
\small
\begin{tabular}{@{}llll@{}}
\toprule
                & $\Delta<\Delta(n)$ & $\Delta>\Delta(n)$ & $\Delta=\Delta(n)$ \\ \midrule
XLM-Twitter     & 36.41    & 6.45     & 59.91       \\
XLM-Politics    & 31.34    & 3.69     & 62.21       \\
GBERT-Sent.& 38.71    & 2.76     & 58.53       \\
GBERT-Mig. & 34.56    & 2.30     & 63.13      \\ \bottomrule
\end{tabular}
\caption{Comparison of computational approaches regarding relative difference values  between PNC and name valence based on PLMs ($\Delta$) and valence norms ($\Delta(n)$). PNC proportions are provided in percent with minimum PNC frequency = 5.}
\label{tab:approach-comparison}
\vspace{-0.4cm}
\end{table}

\begin{table*}[!htpb]
\small
\centering
\begin{tabular}{@{}lllllllll@{}}
\toprule
& \multicolumn{2}{c}{Goal-Klose} & \multicolumn{2}{c}{Prison-Hoeneß} & \multicolumn{2}{c}{Welcome-Merkel} & \multicolumn{2}{l}{Expropriation-Kühnert} \\ \midrule
PLM               & $v$(PNC)  &  $v$(Name) &  $v$(PNC)  & $v$(Name) &  $v$(PNC)  & $v$(Name) &  $v$(PNC)  & $v$(Name) \\ \midrule
XLM-Twitter     & 5.63 & 4.86 & 3.39 & 4.31 & 3.18 & 4.31 & 2.95 & 3.96 \\
XLM-Politics    & 4.38 & 3.38 & 2.58 & 2.37 & 0.00 & 2.35 & 1.10 & 1.80 \\
GBERT-Sent. & 4.58 & 4.82 & 2.74 & 4.74 & 3.64 & 4.89 & 3.18 & 4.63 \\
GBERT-Mig. & 6.04 & 4.64 & 3.23 & 4.36 & 2.73 & 4.44 & 3.18 & 4.25 \\ \bottomrule
\end{tabular}
\caption{Comparison of model predictions for sample PNCs where $v$(PNC) and $v$(Name) denote name and PNC valence on a scale from 0 to 10 with 0 and 10 referring to low and high valence, respectively. }
\label{tab:model-comparison}
\vspace{-0.4cm}
\end{table*}

\paragraph{Domain-Specific Results} Further zooming in on examples (cf. Table~\ref{tab:model-comparison}), we find all models but GBERT-Sentiment predicting the PNC \textit{Tore-Klose} (`Goal-Klose') more positively evaluative than the name Miroslav Klose. The same is true for \textit{Knast-Hoeneß} (`Prison-Hoeneß') where all but XLM-Politics predict the PNC to be more negatively evaluative than the name Ulrich Hoeneß. Inspecting PNCs with very high and low modifier valence, we observe that all models agree on the PNC \textit{Willkommens-Merkel} (`Welcome-Merkel') carrying a more negative meaning than the full name which is line with our valence norms experiment. Similarly, for the PNC \textit{Enteignungs-Kühnert} (`Expropriation-Kühnert') with very low modifier valence, model predictions match regarding a more negatively evaluative PNC compared to the name Kevin Kühnert.

\paragraph{Human Evaluation of PNCs}
We perform a human evaluation of sentiment predictions to assess task difficulty and compare model predictions to humans' opinions. 
For this, we evaluate 30 PNCs (10\% of targets) for which (i)~all PLMs and valence norms predict the same label (negative only), (ii)~PLMs agree among themselves but disagree with valence norm prediction, and (iii)~PLMs disagree among themselves and/or disagree with valence norms prediction. For feasibility reasons, we focus on PNCs occurring within max. 30 contexts (avg. \# contexts: 12.7). Provided a PNC in context, five annotators are asked to annotate sentiment choosing between the labels \{positive, negative, neutral\}. 
The evaluation task is carried out online in a remote setting using Google Forms and Google Tables. We collect unique and complete answer sets from six annotators.\footnote{For further details on the annotators and the annotation setup and, we refer to Sec.~\ref{sec:ethics}.} Pairwise inter-annotator agreement\footnote{We exclude submissions from one annotator due to poor inter-annotator agreement.} ($\diameter \rho=0.61$) indicates reasonable consensus. Self-reported task difficulty reveals that annotators assess the annotation as difficult in at least 40-60\% of the cases noting that they needed time to choose a label and expressing uncertainty in some cases. 
Using the obtained annotations, we determine valence as described in Eq.~\ref{eq:2}.

\begin{figure*}[!htpb]
    \centering
    \includegraphics[width=\textwidth]{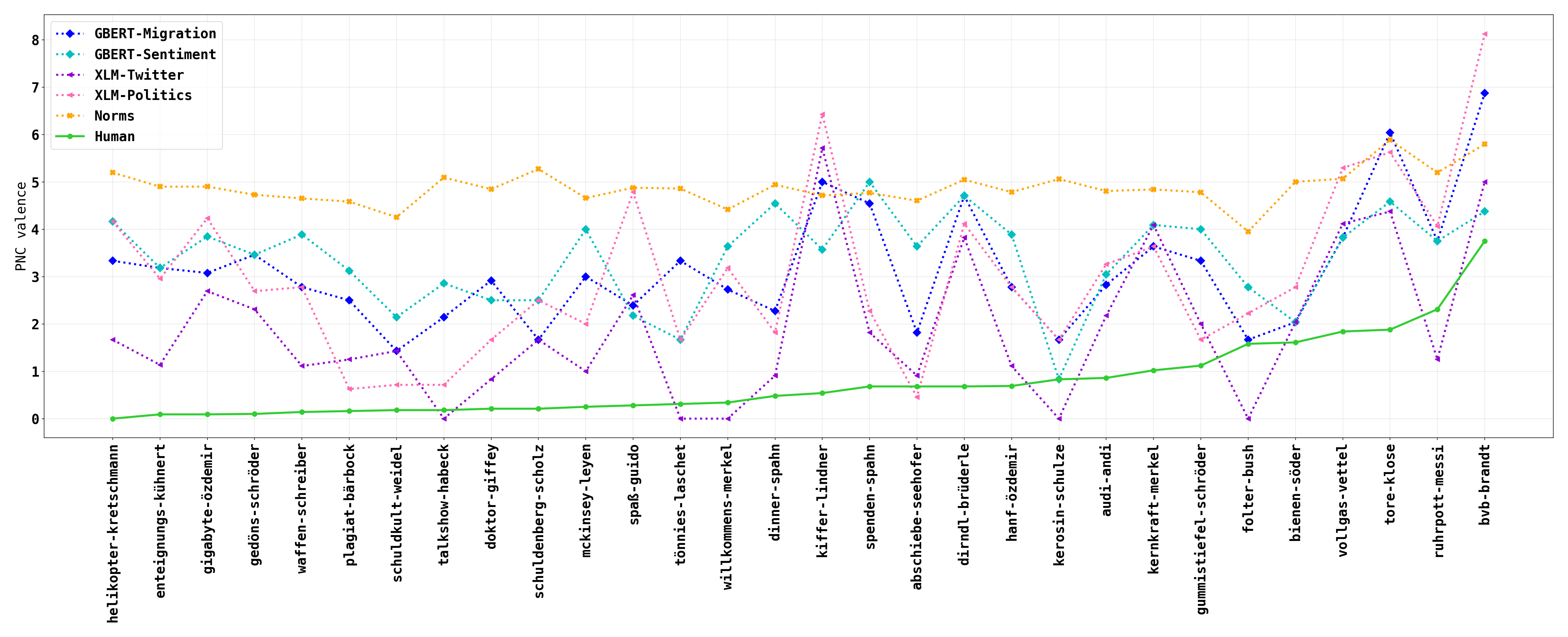}
    \caption{Comparison of PNC valence at discourse level determined by humans (solid green line), valence norms (dashed orange line, x-markers), XLM-based PLMs (dashed pink and violet lines, triangle markers), and German BERT-based PLMs (dashed blue and cyan lines, rhombus markers).} 
    \vspace{-0.2cm}
    \label{fig:models_vs_human}
\end{figure*}

A visualization of PNC valence scores determined by human annotations compared to computational approaches (valence norms, PLMs) is shown in Fig.~\ref{fig:models_vs_human}. While human evaluation suggests a substantially more negative meaning of most PNCs as compared to all computational approaches, domain-specific differences are underlined with all PNCs from the domain \textit{sports} evaluated more positively than PNCs from the domain \textit{politics}.
Furthermore, while PLM predictions have high variance which seems to be connected to modifier meaning, both our valence norms-based approach and human evaluation shows less variance which hints towards stronger incorporation of the whole discourse. These observations point towards the need of further investigation, for example, focusing on predicting sentiment towards a specific target \cite{pei2019targeted} or increasing model attention on the discourse as a whole.
Furthermore, human evaluation seems to be less influenced by modifier meaning in contrast to PLM predictions, e.g., ratings for \textit{Folter-Bush} (`Torture-Bush') and \textit{Bienen-Söder} (`Bee-Söder') are assigned almost equal ratings by humans, while PLM and valence predictions differ quite significantly.

\smallskip

Based on these observations, we confirm that sentiment predictions obtained from PLMs fine-tuned for SA can serve as a proxy for assessing whether PNCs are more negatively or positively evaluative than corresponding personal names. However, we find model-specific differences including (i) PLMs providing stronger valence assessments than our valence-based approach with a tendency towards more negative predictions and (ii) XLM-based models suggesting a more positive interpretation of PNCs than models using German BERT as their backbone.
A comparison of models with human evaluation of PNCs however hints towards a more negatively evaluated meaning of PNCs compared to both computational approaches with PNCs from the domain \textit{sports} evaluated more positively than from the domain \textit{politics}.

\section{Regression Modelling} \label{sec:rm}
Given that every PNC refers to a real-world person, e.g., \textit{Tore-Klose} (`Goal-Klose') $\rightarrow$ \textit{Miroslav Klose}, we hypothesize that personal background information potentially influence the evaluative meaning of a PNC. Furthermore, evaluating PNCs via valence (cf. §\ref{sec:valence}) and PLMs (cf. §\ref{sec:plms}) showed that information such as modifier valence impacts whether a PNC is more positively or negatively evaluated than a full name. To explore which factors are in fact influential at a statistically significant level, we first enrich each name and PNC with 
personal (age, gender, nationality, place of birth), domain-specific (domain, political party membership), and extra-linguistic information (name, PNC, and modifier valence scores, event frame). In a next step, we fit a range of linear regression models to see which relationships are directed and explore options for variable selection.

\subsection{Data}
As valence scores calculated with valence norms (cf. §\ref{sec:valence}) show a moderately positive correlation between name and valence (as compared to PLM-based valence), we draw on corresponding valence scores for this analysis. We use all 289 targets for which a valence score for both the full name and the PNC could be calculated. Then, we determine the corresponding $\Delta$ where positive values refer to cases where compounds are more positive than the name and negative values represent target pairs where the compound is more negative than the name. For each PNC, we also compute the corresponding modifier valence scores and filter out cases where no score could be determined.

\paragraph{Data Enrichment} For each target, we manually collect and assign relevant personal background information based on publicly accessible information\footnote{We use Wikipedia through Google to collect data.}. We further model the relationship between the modifier and the compound head as frame elements of the event frame in which the name bearer participated, using German FrameNet:\\

\begin{itemize}[leftmargin=*]
    \item \textbf{Domain} (politics: 87\%, sports: 9\%, show business: 1\%, others: 3\%)
    \item \textbf{Current age} in full years (if deceased: age at time of death)
    \item \textbf{Current nationality} (Argentina, Austria, France, Turkey: <1\%, Germany: 88\% Russia, Sweden, UK: 1\%, US: 7\%)
    \item \textbf{Place of birth} (for feasibility restricted to West Germany: 77\%, East Germany: 16\%, outside of Germany: 7\%)
    \item \textbf{Gender} (female: 22\%, male: 78\%)\footnote{No target identified as non-binary according to publicly available information.}
    \item \textbf{Political party membership} (Austria: Team HC Strache: <1\%, Germany: AfD: 5\%, CDU/CSU: 25/10\%, FDP: 12\%, The Greens: 12\%, The Left: 1\%, SPD: 18\%, Centre: <1\%; Russia: United Russia: 1\%; UK: Conservatives: <1\%; USA: Democrats: 3\%, Republicans: 2\%; non-party politicians are assigned the label \textit{independent}; people who are neither politicians nor party members are assigned \textit{no party})
    \item \textbf{Participation in events} (20 \href{https://gsw.phil.hhu.de}{German FrameNet} frames; PNCs not representing an event corresponding to a frame or an unknown event are labeled \textit{not eventive} or \textit{unknown}, respectively)
\end{itemize}

\subsection{Linear Regression Modelling}

\paragraph{Univariate Linear Regression} We first fit a linear model to predict $\Delta$ using each of our ten independent variables\footnote{Reference categories for factor variables are determined by lowest value.}
and find significant results for the variables PNC valence, modifier valence, age, political party, and birthplace (cf. App.~\ref{appsubsec:univariate}, Table~\ref{tab:univariate}), 
no significant difference in means based on the Tukey post-hoc test). 

The results indicate PNC valence as highly significant predictor explaining \textasciitilde88\% of variance. Modifier valence also has a positive linear relationship, explaining around 10\% of variance, while age comes with an significant inverse relationship, i.e., increasing age seems to be reflected in a PNC that is more negative than the name of a person itself. If a person is a member of the far right party AfD (Alternative for Germany), a negative $\Delta$ is more likely, while being a member of any other larger party is positively related to $\Delta$ as compared to the AfD. In particular, members of The Greens party are likely to be assigned a positive $\Delta$. Moreover, different birth places might be connected to differences in the evaluative nature of a corresponding PNC.

\paragraph{Variable Selection with Elastic Net}
To further verify which predictors are relevant, we fit a linear regression model using elastic net regularization leveraging all variables excluding either name valence or PNC valence or both variables. Further details are reported in App.~\ref{appsubsec:elasticnet}. 

Excluding PNC valence leads to increased importance of modifier valence as well as personal and domain-specific information such as age, while name valence is of low relevance. In general, the fitted model only explains a limited amount of variance ($R^{2}=0.15, \alpha=0.16, \lambda=0.06$). Excluding both name and PNC valence increases the importance of modifier valence and variables focusing more on geographic information such as place of birth. Similarly to the previous model, this model only explains a limited percentage of variance ($R^{2}=0.15, \alpha=0.005, \lambda=0.26$). Excluding name valence places maximum importance on PNC valence as well as domain categories followed by personal information, while modifier valence only bears little relevance. The fitted model clearly outperforms the other models, explaining a high percentage of variance ($R^{2}=0.92, \alpha=0.71, \lambda=0.001$). 

\paragraph{Multivariate Linear Regression}
As a next step, we fit a range of multivariate regression models based on theoretical background, including models based on (i) personal information including age, gender or age, gender, nationality, origin, (ii) and domain-specific information such as domain and political party membership, as well as (iii) semantic knowledge and extra-linguistic information regarding the PNC encompassing modifier valence, FrameNet (and PNC) valence. \footnote{As $\Delta$ is calculated based on name and PNC valence, including both would yield a model of perfect fit, which does, however not reveal potentially interesting results. We thus only consider scenarios where either one or both variables are excluded.}
Additionally, (iv) three models including all but either name or PNC valence or neither of both variables are fitted.

Results (cf. App.~\ref{appsec:lr}, Table~\ref{tab:multivariate}) reveal that only models based on (i) personal, or (ii) domain-specific information are significant but cannot explain the variance in the data very well (best model \texttt{\small delta\textasciitilde party + domain} yields $Adj. R^{2}=0.11, p<2.2*10^{-16}$). Models based on (iii) extra-linguistic information regarding the compound, on the other hand, are more successful with adding PNC valence significantly boosting performance ($Adj. R^{2}=0.89, p<2.2*10^{-16}$). 
(iv) Including all variables except either name or PNC valence or neither of both yields models of mixed performance with PNC valence excluded leading to significantly less variance explained (PNC excluded: $Adj. R^{2}=0.12, p<0.01$, name and PNC valence excluded: $Adj. R^{2}=0.26, p<0.01$). The best model includes all variables except name valence ($Adj. R^{2}=0.96, p<2.2*10^{-16}$). 

Overall, we observe a highly significant positive relationship for PNC valence. 
Interestingly, we find the domain \textit{sports} connected to a slightly inverse relationship with top valence scores for athlete's names and not the PNC. An example is the PNC \textit{Gold-Rosi} (`Gold-Rosi') where top valence scores are related to the full name \textit{Rosi Mittermaier} and PNC valence scores placed slightly below. In this case, the reason is hidden in many of the contexts who are -- very positive, but nevertheless -- obituaries of the famous skier (cf. Table~\ref{tab:data-example}, Fig.~\ref{fig:pnc-name})
If a person identifies as male or comes from the U.S., the PNC is slightly more likely to be more positively or negatively evaluative, respectively. Political party membership has an inverse relationship in cases of the CDU, CSU, FDP, The Greens, and the SPD, while Democrats and Republicans come with a positive relationship (reference level: AfD). 

\section{Conclusion}
We tackled the under-studied task of modeling the meaning of PNCs such as \textit{Willkommens-Merkel} ('Welcome-Merkel'), and presented a comprehensive computational exploration revealing that PNCs are both positively and negatively evaluative at discourse level. We examined 321 German PNCs from domains such as politics and sports and their respective full names, e.g., Angela Merkel. We developed two computational approaches based on (i)~valence norms and (ii)~PLMs and compared results to human annotation, uncovering domain-specific differences where athletes are generally evaluated more positively than politicians. To explore PNC connections to the respective real-world persons, we enriched our data with personal background information and employed regression analyses to demonstrate which factors influence PNC valence. 

\section*{Limitations}
In this work, we concentrate on PNCs in German. As far as the transfer of the suggested approach to languages other than German is concerned, we call attention to the potential need for valence norms in a specific language that might not be readily available in a specific language. Researchers could draw on the approach presented in \cite{koper-schulte-im-walde-2016-automatically} to automatically generate valence norms in the desired language. Since it might however be difficult to find a sufficient amount of written text in the case of some languages, we present an approach using PLMs to obtain valence assessments using sentiment predictions as a proxy. While multilingual PLMs support a great range of languages, a specific language might not be included or under-represented in the training data. In these cases, adapter-based approaches \cite{pmlr-v97-houlsby19a,pfeiffer-etal-2022-lifting} requiring limited amounts of text might be an alternative to obtain sentiment predictions in a desired language.

To explore PNCs at discourse-level, we use Twitter and news text. Since we only obtain 100 tweets per full name, a systematic comparison between PNCs in Twitter vs. news text is not possible. Future work could however investigate  whether social media networks who can be used by anyone foster or change the use and function of PNCs as compared to news text authored by professionals and mainly intended to provide a unilateral information flow to the reader without expecting immediate reaction or discussion. 

In our work, we leverage the Leipzig Corpora collection that provides large amounts of German news data in the context of the ongoing project Deutscher Wortschatz (DW) \cite{klein-geyken-2010,goldhahn-etal-2012-building}, spanning a time period of 27 years (1995-2022). In a pilot study, we also experimented with the \href{https://commoncrawl.org/blog/news-dataset-available}{Common Crawl News Dataset}. We compared the results with those from DW, however, there was only a small gain in collected contexts which did not justify the effort of processing several terabytes. We therefore decided to use DW to save resources. Another alternative could have been \href{https://commoncrawl.org}{Common Crawl} itself which, however, was not selected because it requires a lot of effort to control the quality of the data. Finally, resources customized to German such as the \href{https://german-nlp-group.github.io/projects/gc4-corpus.html}{GC4 dataset} could be taken into consideration, however, in this case, we would have lost a substantial amount of data since the GC4 spans only 5 vs. the considered 27 years of data.

We would like to mention that the Academic Twitter API was unfortunately closed and can only be leveraged through a paid API to to re-create this part of the used context corpus. In contrast, the subcorpus we built using DW is fully reproducible.

\section*{Ethics Statement} \label{sec:ethics}

We leverage PLMs as provided and licensed under the Apache License 2.0 by \texttt{huggingface} \cite{wolf-etal-2020-transformers}. We acknowledge that valence assessments predicted using the outlined approach are a product of unsupervised learning methods which might be prone to error. We point out that predictions should be approved by an expert or flagged otherwise in case they are used in a downstream application to avoid potential risks such as biased decisions.

In the context of our evaluation task, we collected sentiment ratings from human participants. For this, the participants were provided an informed consent declaration with the name and the contact of the principal investigators; the title, purpose and procedure of the study; risks, benefits and compensation for participating in the study; confirmation of confidential anonymous data handling; and confirmation that participation in the study is voluntary. The informed consent declaration was signed by the participants before taking part in the study. 
Annotators were provided written guidelines including example questions and borderline decisions. In case of questions, annotators had the option to contact the authors of the paper. The evaluation task was carried out online in a remote setting using Google Forms and Google Tables. The annotation task was completed by one author and five externally recruited annotators who have no connection to any of the authors' affiliations. External annotators received compensation according to our country's minimum wage regulations for their effort. All annotators are native speakers of German. 
The evaluation could be completed flexibly within four days. Annotators could take as much time as needed to complete the evaluation (average time effort: \textasciitilde1.15 hour).
Each annotator submitted one unique set of answers. 

\section*{Acknowledgements}
We are grateful to the IMS SemRel group for helpful suggestions and feedback regarding this work. We would also like to thank the anonymous reviewers for their comments and suggestions. 

Annerose Eichel was funded by the Hanns Seidel Foundation's Talent Program. Sabine Arndt-Lappe (2019-2023) and Milena Belosevic (2019-2022) received funding from a grant by the Forschungsinitiative Rheinland Pfalz 2019-2023, Verbundprojekt \textit{Patterns}, Linguistic Creativity and Variation in Synchrony and Diachrony. This research was further supported by the DFG Research Grant SCHU 2580/5-1 (Computational Models of the Emergence and Diachronic Change of Multi-Word Expression Meanings). 

\section{Bibliographical References}\label{sec:reference}

\bibliographystyle{lrec-coling2024-natbib}
\bibliography{lrec-coling2024-example}


\appendix
\section{Data} \label{appsec:data}

\paragraph{Search Heuristics} To find the maximum possible number of sentences that contain a PNC, the PNC list was modified at character level. We duplicate our PNC list and apply the following heuristics:\\
\begin{itemize}[leftmargin=*]
\itemsep +2mm
    \item \textbf{Umlauts:} Replace umlauts: ä $\rightarrow$ ae, ö $\rightarrow$ oe, ü $\rightarrow$ ue, e.g., \textit{Bätschi-Nahles} $\rightarrow$ \textit{Baetschi-Nahles} (`Bätschi\footnote{`Bätschi' is typically used by children to express mischievous mockery (often combined with a special gesture), demonstrating that oneself owns, knows, or feels more or better than the other person.}-Nahles')
    \item \textbf{Eszett:} Replace ß $\rightarrow$ ss, e.g., \textit{Spaß-Guido} $\rightarrow$ \textit{Spass-Guido} (`Fun-Guido')
    \item \textbf{Interfix:} Add or delete the interfix accordingly, e.g., \textit{Hoffnungs-Obama} $\rightarrow$ 
    \textit{Hoffnung-Obama} (`Hope-Obama')
    \item \textbf{Alternative spelling:} Included spelling variations of words, e.g., \textit{Gazprom-Schröder} $\rightarrow$ \textit{Gasprom-Schröder} (`Gasprom-Schröder')
    \item \textbf{Singular/Plural:} Added or deleted a letter to get the singular/plural form of the modifier, e.g., \textit{Tore-Klose} $\rightarrow$ \textit{Tor-Klose} (`Goal-Klose')
    \item \textbf{Wildcard search:} Added a wildcard (limited to 0-2 characters) between modifier and head to find PNCs without a hyphen/with a space/with a hyphen and hashtag/etc. inbetween.
\end{itemize}

\paragraph{PoS Tags for Valence Exploration} 
All context words of a target are tagged with POS labels using the  \href{https://www.cis.uni-muenchen.de/~schmid/tools/TreeTagger/}{probabilistic TreeTagger \cite{Schmid1999}}. In case of unknown lemmas, the word itself is used to avoid losing context data. The context words are then filtered to exclude words such as determiners, prepositions, pronouns, modal verbs, punctuation, etc. of which the valence value has little interpretable meaning. 

\smallskip

{Included PoS Tags:}
\vspace{+1mm}
\begin{itemize}
\itemsep +1mm
    \item \textbf{NN}: simple noun
    \item \textbf{ADJA}: attributive adjective
    \item \textbf{ADJD}: predicative or adverbial adjective
    \item \textbf{VVFIN}: finite full verb 
    \item \textbf{VVIMP}: imperative (full verb) 
    \item \textbf{VVINF}: infinitive (full verb) 
    \item \textbf{VVIZU}: infinitive with incorporated “zu” particle (full verb) 
    \item \textbf{VVPP}: past participle (full verb) 
\end{itemize}

\section{Regression Analysis} \label{appsec:lr}

All linear regression models are fitted using \texttt{R} \cite{R2023}.

\subsection{Univariate Regression Modeling} \label{appsubsec:univariate}

To explore which single predictors are most relevant, we fit a linear regression model for each of our 10 predictors to predict $\Delta$ using the \texttt{lm} package \cite{R2023}.

\begin{table}[!htpb]
\small
\centering
\begin{tabular}{@{}lrrl@{}}
\toprule
\textbf{Predictor} & \textbf{Intercept} & \textbf{Slope} & \textbf{$ (Adj) R^{2}$} \\ \midrule
Name valence      & 0.61  & -0.12  & 0.00  \\ \midrule
PNC valence  & -4.35 & 0.90  & 0.88*** \\ \midrule
Modifier valence  & -0.37 & 0.09   & 0.10***  \\ \midrule
Age               & 0.24  & -0.00 & 0.02**   \\ \midrule
Gender             &  &        &  0.00  \\ \midrule
Domain        &  &        & 0.01  \\ \midrule
Political Party   &   &        & 0.05*   \\
- AfD       &   -0.22  &  & \\
- CDU         &        &  0.63 & \\
- CSU                           &  & 0.25 & \\
- FDP                            &  & 0.31&  \\
- The Greens     &       &    0.42   &\\
- No party        &       & 0.25       &   \\
- The Left        &       &    0.59    &    \\
- Independent & & 0.30 & \\
- SPD             &       &   0.25     &   \\ \midrule
Nationality       &   &        & 0.02   \\ \midrule
Place of Birth            & -0.11 &   & 0.01  \\
- West Germany    &   &  0.15     &    \\ \midrule
FrameNet          &   &    & 0.04   \\ \bottomrule
\end{tabular}
\caption{Univariate regression results separated by horizontal lines with {\footnotesize *$p<0.05$; **$p<0.01$ ***$p<0.001$}. In case of multi-level variables, adjusted $R^{2}$ is reported.}
\label{tab:univariate}
\vspace{-0.1cm}
\end{table}

We show an overview of univariate linear regression modeling results in Table~\ref{tab:univariate} with $\Delta$ predicted using each predictor separately, e.g., delta \textasciitilde~age. Results are separated by horizontal lines. For readability, we summarize multi-level variable results whenever various levels yield no significant results, e.g., for FrameNet or nationality. In case of significant results, we report only relevant levels, e.g., in case of birthplace only results for places of birth in West Germany as well as the reference level non-Germany are shown.

\subsection{Variable Selection with Elastic Net} \label{appsubsec:elasticnet}

To further explore which predictors are most relevant, we fit three linear regression models using Elastic Net regression. Our goal is to predict $\Delta$ leveraging all variables but either name valence, PNC valence, or both excluded.
All models are fitted using the \texttt{glmnet} package \cite{friedman2010,tay2023}. Data is first centered and scaled. We then search for the best model using 5x5 cross-fold validation with random search and a tuning length of 25.

\subsection{Multivariate Regression Modeling} \label{appsubsec:multivariate} 

In the next step, we fit a range of multivariate regression models based on theoretical background, including models based on (i) personal information, (ii) and domain-specific information, and (iii) semantic knowledge and extra-linguistic information regarding the PNC. 
Additionally, three models including all but either name or PNC valence or neither of both variables are fitted (iv-vi). We thus fit five regression models using the \texttt{lm} package \cite{R2023}. 

\begin{table}[!htpb]
\centering
\small
\begin{tabular}{llr}
\toprule
\textbf{Model }                                          & \textbf{$ Adj. R^{2}$}  & $SE$   \\ \midrule
\textbf{(i) Personal information}                            &         &      \\
- Age, gender                                    & 0.02*   & 0.39 \\
- Age, gender, nationality, birthplace  & 0.04*   & 0.40  \\ \midrule
\textbf{(ii) Compound information}                            &         &      \\
- Modifier, FrameNet                      & 0.12*** & 0.38 \\
- Modifier, FrameNet, compound & 0.89*** & 0.13 \\ \midrule
\textbf{(iii) Domain-specific information}                    &         &      \\
- Profession, political party                     & 0.05*   & 0.40  \\ \midrule
\textbf{(iv) Exclude name valence }                           &         &      \\
- All remaining predictors                                  & \textbf{0.96***} & 0.09 \\ \midrule
\textbf{(v) Exclude compound valence }                       &         &      \\
- All remaining predictors                                 & 0.12**  & 0.38 \\ \midrule
\textbf{(vi) Exclude both}                                    &         &      \\
- All remaining predictors                              & 0.11**  & 0.38 \\ \bottomrule
\end{tabular}
\caption{Overview of multivariate regression modeling results, separated by horizontal lines with {\footnotesize *$p<0.05$; **$p<0.01$ ***$p<0.001$}.}
\label{tab:multivariate}
\vspace{-0.3cm}
\end{table}

Table~\ref{tab:univariate} presents regression results using multiple predictor variables with the best-performing model including all predictors but name valence (iv).

\end{document}